\documentclass{ieeeaccess}
\usepackage{cite}
\usepackage{amsmath,amssymb,amsfonts}
\usepackage{algorithmic}
\usepackage{graphicx}
\usepackage{textcomp}
\usepackage{stfloats}
\usepackage{algorithm}
\usepackage{array}
\usepackage{url}
\usepackage{verbatim}
\usepackage[caption=false,font=normalsize,labelfont=sf,textfont=sf]{subfig}

\usepackage{bm}
\usepackage{tabularx}
\makeatletter
\AtBeginDocument{\DeclareMathVersion{bold}
\SetSymbolFont{operators}{bold}{T1}{times}{b}{n}
\SetSymbolFont{NewLetters}{bold}{T1}{times}{b}{it}
\SetMathAlphabet{\mathrm}{bold}{T1}{times}{b}{n}
\SetMathAlphabet{\mathit}{bold}{T1}{times}{b}{it}
\SetMathAlphabet{\mathbf}{bold}{T1}{times}{b}{n}
\SetMathAlphabet{\mathtt}{bold}{OT1}{pcr}{b}{n}
\SetSymbolFont{symbols}{bold}{OMS}{cmsy}{b}{n}
\renewcommand\boldmath{\@nomath\boldmath\mathversion{bold}}}
\makeatother

\def\BibTeX{{\rm B\kern-.05em{\sc i\kern-.025em b}\kern-.08em
    T\kern-.1667em\lower.7ex\hbox{E}\kern-.125emX}}

\begin{document}
\history{Date of publication 10/17/2025, date of current version 10/17/2025.}
\doi{10.1109/ACCESS.2025.3608951}

\title{Is My Data in Your AI? Membership Inference Test (MINT) applied to Face Biometrics}
\author{\uppercase{Daniel DeAlcala}\authorrefmark{1},
\uppercase{Aythami Morales}\authorrefmark{1}, 
\uppercase{Julian Fierrez}\authorrefmark{1}, 
\uppercase{Gonzalo Mancera}\authorrefmark{1}, 
\uppercase{Ruben Tolosana}\authorrefmark{1}, 
and \uppercase{Javier Ortega-Garcia}\authorrefmark{1},
}
\address[1]{Universidad Autonoma de Madrid, C. Francisco Tomás y Valiente, 11, 28049 Madrid, Spain. (e-mail: daniel.dealcala@uam.es, aythami.morales@uam.es, julian.fierrez@uam.es, gonzalo.mancera@uam.es, ruben.tolosana@uam.es and javier.ortega@uam.es)}

\tfootnote{This work has been supported by projects BBforTAI (PID2021-127641OB-I00 MICINN/FEDER), HumanCAIC (TED2021-131787B-I00 MICINN), M2RAI (PID2024-160053OB-I00 MICIU/FEDER) and Cátedra ENIA UAM-VERIDAS (NextGenerationEU PRTR TSI-100927-2023-2). Work conducted in the ELLIS Unit Madrid. D. DeAlcala is supported by a FPU Fellowship (FPU21/05785).}

\markboth
{Daniel DeAlcala \headeretal: Is my Data in your AI? MINT Applied to Face Biometrics}
{Daniel DeAlcala \headeretal: Is my Data in your AI? MINT Applied to Face Biometrics}

\corresp{Corresponding author: Daniel DeAlcala (e-mail: daniel.dealcala@uam.es).}

\begin{abstract}
This article introduces the Membership Inference Test (MINT), a novel approach that aims to empirically assess if given data was used during the training of AI/ML models. Specifically, we propose two MINT architectures designed to learn the distinct activation patterns that emerge when an Audited Model is exposed to data used during its training process. These architectures are based on Multilayer Perceptrons (MLPs) and Convolutional Neural Networks (CNNs). The experimental framework focuses on the challenging task of Face Recognition, considering three state-of-the-art Face Recognition systems. Experiments are carried out using six publicly available databases, comprising over 22 million face images in total. Different experimental scenarios are considered depending on the context of the AI model to test. Our proposed MINT approach achieves promising results, with up to 90\% accuracy, indicating the potential to recognize if an AI model has been trained with specific data. The proposed MINT approach can serve to enforce privacy and fairness in several AI applications, e.g., revealing if sensitive or private data was used for training or tuning Large Language Models (LLMs).
\end{abstract}

\begin{keywords}
Audit, Fairness, Face Recognition, Membership Inference, MIA, MINT, Reliability.
\end{keywords}

\titlepgskip=-21pt

\maketitle

\section{Introduction}
\label{sec:intro}
\PARstart{I}{n} June 2023, the European Parliament adopted its position with respect to Artificial Intelligence (AI) \cite{madiega2021artificial}. Concretely, they have imposed some obligations to all AI companies to guarantee the protection of the citizen rights. For example, this new regulation imposes the registration of AI/ML models in a European Union (EU) database and gives national authorities of EU countries the power to request access to both the trained AI models and the associated procedural details conducted in the development of those models. This new regulation is a game changer, imposing more transparency in the development of AI technologies in Europe. As a result, novel auditing tools based on the access to trained AI models must be developed to monitor AI technologies and their correct deployment in our society. This trend is not limited to the European level, as the White House \cite{USA} has also taken a stance by publishing a memorandum that declares AI a matter of national security, emphasizing the need for access to AI technologies to review them and ensure the protection of citizens' rights.

\begin{figure}[t]
\centering
\includegraphics[width=0.55\columnwidth]{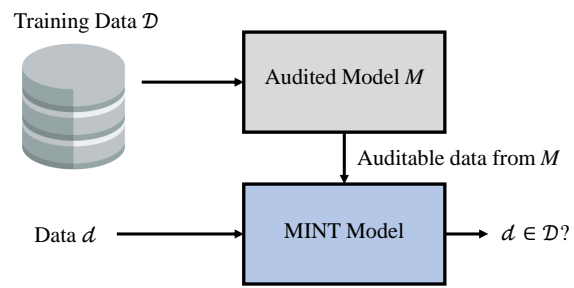} 
\caption{The proposed Membership Inference Test (MINT) model is trained to predict if a given data ($d$) was used during the training process of an AI Model ($M$), trained with a database ($D$).}
\label{Block_diagram_lite}
\end{figure}

Nowadays, AI systems have become increasingly advanced and pervasive across different domains, collecting, analyzing, and processing vast amounts of data. These data often contain sensitive information \cite{2021_TPAMI_SensitiveNets_Morales} about individuals as well as copyright content \cite{Shasha2024Insights}, raising concerns about privacy \cite{Wang2018Inference,Ahmad2022Differential,Marta2017Privacy} and unauthorized access \cite{manheim2019artificial, pena2023human, delgado2022survey}. Consequently, this necessitates the fast development of novel tools to clarify what data has been used to train AI models, preventing therefore the use of non-authorized data and increasing the transparency and explainability of the AI models \cite{Rong2024Towards, BARREDOARRIETA202082, 2023_ECAI}. These tools aim to prevent learning frameworks from hiding the training data within the model's parameters, thereby unveiling the opacity that they have traditionally relied upon. These considerations lead us to the main objective of the present study: checking if given data was used to train AI models \cite{Jegorova2023Leakage}, as summarized in Fig. \ref{Block_diagram_lite}.

This objective is related to the field of Membership Inference Attacks (MIAs), privacy attacks that target AI models trained on sensitive data \cite{shokri2017membership,carlini2022membership}. MIAs seeks to infer membership information about the training dataset used by exploiting the behavior and responses of the model. Also, MIAs leverage the potential memorization of the training data used by the AI models, especially when the models overfit \cite{salem2018ml,leino2020stolen} or when the training data contain unique or distinctive patterns \cite{shokri2017membership}. However, recent studies question MIAs effectiveness due to the inherent complexity of the task \cite{rezaei2021difficulty,hu2022membership}. 

While MIAs are traditionally considered as attacks to quantify the potential privacy leakage of AI models, in the present article we introduce a novel perspective in line with the new AI regulation \cite{madiega2021artificial, USA}. Concretely, we consider it as an auditing tool for detecting the potential use of unauthorized data for training AI models, a concept we term as Membership Inference Test (MINT). The aim of MIAs and MINT is similar: to determine whether a given data sample was used during the training of a target AI model. However, although MIAs and our proposed MINT share the same objective, they diverge in their environmental conditions, leading to different methodological approaches. In MIAs, the model is trained assuming the role of an attacker who does not have access to the original model, but to a replica as similar as possible to it (shadow model) \cite{shokri2017membership}. In MINT, we assume the role of an auditor who has direct access to the original model or partial information of it (e.g., intermediate activations). This is aligned with the new AI regulation from USA \cite{USA} and EU \cite{madiega2021artificial}, which imposes the registration of the trained AI models in an EU database. As a result, this new paradigm moves from the attacker to the auditor point of view, allowing the development of MINT models under different environmental conditions.


In the present study we concentrate on the challenging task of Face Recognition, due to the sensitive nature of face biometrics data \cite{Yueming20233D,Melzi2023WACV,Serna2022Sensitive} according to the General Data Protection Regulation (GDPR) promoted by EU \cite{GDPR}, the California Consumer Privacy Act (CCPA) \cite{CCPA}, and the Biometric Information Privacy Act (BIPA) \cite{BIPA}. Although MIAs and MINT differ due to their distinct environmental conditions, they also share significant similarities, so methods in one domain can inspire advancements in the other as well. The application of MIAs in Face Recognition remains underexplored, with most studies focusing on simple models featuring limited convolutional layers and designed for classification \cite{li2021membership,song2019privacy,yang2020defending,NIU2024404}. These works do not address state-of-the-art models used in Face Recognition, leaving a significant gap in the literature. This gap in the state of the art is particularly crucial as MIAs frequently deal with classification tasks, using the network's output vector to determine membership in the training data, like a form of `probability' of belonging to each class \cite{hu2022membership, choquette2021label, pyrgelis2017knock,truex2019demystifying,salem2018ml}. However, due to the characteristics of Face Recognition systems, focused on recognition rather than classification, such vector usually does not exist. This multiclass probability vector is traditionally replaced by an embedding feature vector representing individual characteristics in a learned space. Consequently, the methods and results diverge even more from what we typically see in state-of-the-art MIAs studies \cite{hu2022membership}.

Some recent studies have explored the application of Membership Inference Attacks (MIAs) on realistic Face Recognition models (\cite{Min2023Face, li2022user, hintersdorf2022does}). However, these works focus on determining whether an entire user was included in the training set, leveraging specific model characteristics. For instance, \cite{Min2023Face, li2022user} demonstrated that users present in the training set tend to have their embedding feature vectors closer to each other. Similarly, \cite{hintersdorf2022does} utilized guided captioning on CLIP models to detect the presence of users in the training set.

The detection of users in a model’s training set, while interesting, has notable limitations compared to the detection of specific samples used in training, which is the focus of MINT. User detection cannot pinpoint which specific samples were utilized during training; for instance, even if consent is provided for certain images, it does not mean all images of a person can be used. Sample detection, as addressed in this work, resolves this issue by identifying precisely which samples were included. Moreover, sample detection inherently achieves user detection, but the reverse does not hold, making the approach in this work more precise and granular. Additionally, user detection methods are restricted to face recognition models, whereas the methodologies proposed in this work are generalizable to any type of model. In fact, similar ideas are applied to other types of models in Appendix A.

\noindent The main contributions can be summarized as follows:

\begin{itemize}
    \item We introduce MINT as a novel perspective to detect if given data was used in an AI model’s training from an AI auditor standpoint.
    \item We propose two MINT architectures: \textit{i)} the Vanilla MINT model based on a Multilayer Perceptron (MLP) network, trained by applying max pooling to the activation maps of both training and non-training data samples, and \textit{ii)} the Convolutional Neural Network (CNN) MINT model, trained using the entire activation blocks of both training and non-training data samples. 
    \item We conduct an extensive evaluation of the two proposed MINT architectures considering three state-of-the-art Face Recognition systems and six publicly available databases, comprising over 22 million face images. Different experimental scenarios are considered depending on the context available of the Face Recognition systems. The proposed MINT approaches achieve promising results, with up to 90\% accuracy detecting if data was used for training, reducing the error rates by over 55\% compared to other state-of-the-art methods.
    
\end{itemize}

The article is structured as follows. Sec. \ref{sec:Related Works} provides a revision of the state of the art in the scope of the article. Sec. \ref{sec:MINT} describes the main concepts of the proposed MINT, and the specific details for the task considered, Face Recognition. The databases, Face Recognition systems, and experimental protocol are described in Sec. \ref{ExperimentalFramework}.  Results are discussed in Sec. \ref{Evaluation}, whereas the significance of Face Recognition in MINT and the challenges of developing this technology in real-world applications are discussed in Sec. \ref{sec:whyface} and \ref{sec:realworld}.
Discussion and conclusions are finally drawn in Sec. \ref{Discussion} and \ref{Conclusions}. 

\section{Related Works}
\label{sec:Related Works}

First, it is important to remark that, to the best of our knowledge, the specific problem addressed in the present article (MINT) has not been directly tackled in the current literature. Nevertheless, there are studies and research directions that touch upon similar topics or are grounded in analogous principles, which we will briefly discuss in this section. Our study closely relates to three fundamental research directions: ``Membership Inference Attacks'', ``Reconstructing Training Data from Neural Networks'', and ``Data Tracing''. In the following, we provide concise explanations of these three primary research areas, with special emphasis on MIAs due to its significant similarity to our own research.

\subsection{Membership Inference Attacks}

The capability to detect the data used for training AI models presents a security concern that can potentially unveil sensitive \cite{2021_TPAMI_SensitiveNets_Morales} or private information \cite{melzi2022overview}, encompassing areas such as shopping preferences, health records, and pictures, among many others. This capability has been explored in the literature mainly under the name of Membership Inference Attacks (MIAs). These attacks aim to extract such sensitive information through an adversarial approach, without access to the model or its training data, as providers are hesitant to disclose such data. The pioneering work by Shokri \textit{et al.}  \cite{shokri2017membership} laid the groundwork for this line of research. In their approach, they used ``shadow models'' that emulate the functionality of the original model, despite having no direct access to it. To construct these so-called ``shadow models'', it is essential to have knowledge about the architecture of the original model they are based on, along with a subset of samples and statistics derived from the original dataset used for training. As a result, they crafted ``shadow training sets'' mirroring the original dataset, subsequently training shadow models to mimic the original model. This approach offers complete control over the training and non-training data of these models. A binary classifier was trained to distinguish between the training and non-training data used in the shadow models. Images were passed through the network, and output embeddings from all shadow models were used to train a simple binary classifier. Better results were observed with an increasing number of shadow models.

Other approaches have emerged from the original publication \cite{shokri2017membership}, following the same principle, but instead of training a binary classifier, relying on the value of specific metrics and thresholds. For example, the authors of \cite{yeom2018privacy} analyzed whether the loss value of input data was above a threshold or not, or \cite{song2021systematic,salem2018ml}, which used the prediction value.

An alternative method was presented by Nasr \textit{et al.} \cite{nasr2019comprehensive}. Their work introduced the Black-box and White-box terminology into this context. Previous efforts primarily relied on the output of shadow models for classification, essentially the output embedding (Black-box). However, Nasr \textit{et al.}  proposed a White-box framework, granting them access to the activations, loss, and gradients. The authors demonstrated that access to the White-box information offers limited utility in distinguishing whether a sample was used for training or not. The best results were achieved using gradients \cite{nasr2019comprehensive} where as activations did not yield promising results \cite{rezaei2021difficulty,cretu2023re}. In general, in the White-box context, the number of studies in the literature is limited, and with no significant results.

More recently, the authors of \cite{rezaei2021difficulty} delved into the challenges of MIAs and the inherent complexity of the task. They concluded that the actual performance in the task is notably worse than the results published in the state of the art. This is largely due to the insufficient evaluation methodology. Furthermore, they explored previously untapped White-box information involving gradients, although they were not able to significantly improve the Black-box framework performance. 
Shafran \textit{et al.} \cite{shafran2021membership} analyzed how factors such as task complexity and image resolution impact the results of MINT.


\subsection{Reconstructing Training Data from AI Models}
The goal of MIAs is to figure out the data used to train the AI model with some prior knowledge of the data. On the contrary, the research line in the present section is purely based on the network itself, without any prior knowledge of the data. Initial methods tried to find inputs that made intermediate neurons highly active. However, this often resulted in noisy data \cite{erhan2009visualizing}. To tackle the noisy data problem, some approaches introduced prior knowledge or image generators \cite{mahendran2015understanding} \cite{yosinski2015understanding}. The drawback is that the image generators can lead to adversarial examples \cite{goodfellow2014explaining,Mahdi2023Towards}. The recent work by Haim \textit{et al.}  \cite{haim2022reconstructing} represents a cutting-edge approach, showing very promising outcomes. The techniques described there can generate images resembling those used for training, offering insights into the utilized images, but they can not guarantee if specific images were part of the training set or not. It is worth noting other similar approaches like model-inversion, which aims to find representatives of each output class \cite{fredrikson2015model}, \cite{yin2020dreaming}.

\subsection{Data Tracing}

Methods described in previous sections can be seen as ``passive methods'' that rely on an already trained model without interfering with it. In contrast, the Data Tracing research direction falls under the category of ``active methods'', where the training data is directly altered to influence the model \cite{Goldblum2023Dataset}. Data tracing involves changing the training data to identify if it was used during training. Several studies have tried to tackle this challenge, demonstrating that with smart changes in the training data, it is feasible to detect its presence during inference \cite{biggio2012poisoning,steinhardt2017certified}. For instance, Radioactive Data \cite{sablayrolles20a} achieved this by inserting ``radioactive marks'' into the training data to detect if a database was used to train a model.

\begin{figure}[t]
\centering
\includegraphics[width=0.80\linewidth]{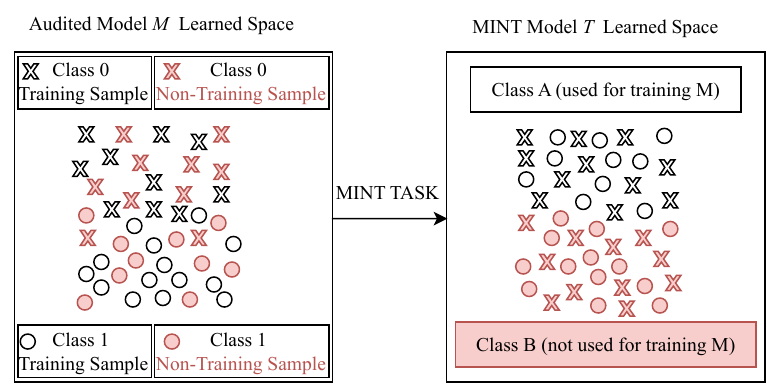} 
\caption{The MINT task is represented graphically. On the left, we show the Audited Model Learned Space, illustrating two classes (0 and 1) each with samples used and not used in the Audited Model training. On the right, we present the feature space we aim to learn with our proposed MINT Model, where previous embeddings represented in the left plot will be projected so they become easily separable for the new binary classification task: used or not used (A or B) in the training of the Audited Model. }
\label{MINTTaskSimple}
\end{figure}

\section{Membership Inference Test (MINT): Proposed Approach} \label{sec:MINT}

MINT is a Membership Inference Test for a neural network's training data that aims to comply with current legislation \cite{madiega2021artificial, USA}, hence, assuming a certain level of collaboration from the model developer. The ultimate goal is to apply the MINT techniques to all data types (text, images, audio, etc.), considering varying levels of information provided by the model developer about the model training and operation. This spans from scenarios where no data information is provided (unsupervised) or very little (semi-supervised), up to when sufficient information about the training data is granted (supervised). Simultaneously, it considers whether you have access to the entire model (white-box) or merely to the model's output (black-box). In the present study, we analyze a broad range of scenarios, varying the quantity of data available for training the MINT model and contemplating both white-box and black-box settings.

Detecting the data used to train a model is a complex task in some sense opposite to the primary goal of common machine learning (ML) problems: generalize well during inference to samples unseen during learning, i.e., behaving similarly for both training and non-training samples. Consequently, in standard ML we expect samples of the same class to be proximate in the model representation space, regardless of their presence in the training set (Fig. \ref{MINTTaskSimple}). MIAs and MINT on the other hand, seek to distinguish samples (used or not for training the Audited Model $M$) that the model $M$ intends to represent as close as possible. Hence, much of the existing MIA literature revolves around overfitted networks that do not generalize well \cite{shokri2017membership,tonni2020data,irolla2019demystifying,hu2022membership,hu2021ear,kaya2021does}. In this work, however, we apply MINT to state-of-the-art FR models, rather than overfitted models, to evaluate the real applicability of our method.

In MINT, we assume the role of model auditors with access to the original model, or at least to partial information of it. This is a consistent scenario either due to the legal considerations outlined in the present study (i.e., granting national authorities the possibility to request the model \cite{madiega2021artificial, USA}), or as a result of developer transparency initiatives aimed at end-users. This is an important distinction from related works around MIAs, as here we do not need to train ``shadow models'' to simulate the original model's behavior. The ``shadow models'', while a useful approximation, operate in a space different from the original (Fig. \ref{MINTTaskSimple} left), potentially imposing limitations on the achievable performance \cite{rezaei2021difficulty}. Instead, with access to the original model, we can directly apply our proposed approach, which may yield different detection results and methods for similar tasks.

\begin{figure*}[t]
\centering
\includegraphics[width=0.95\linewidth]{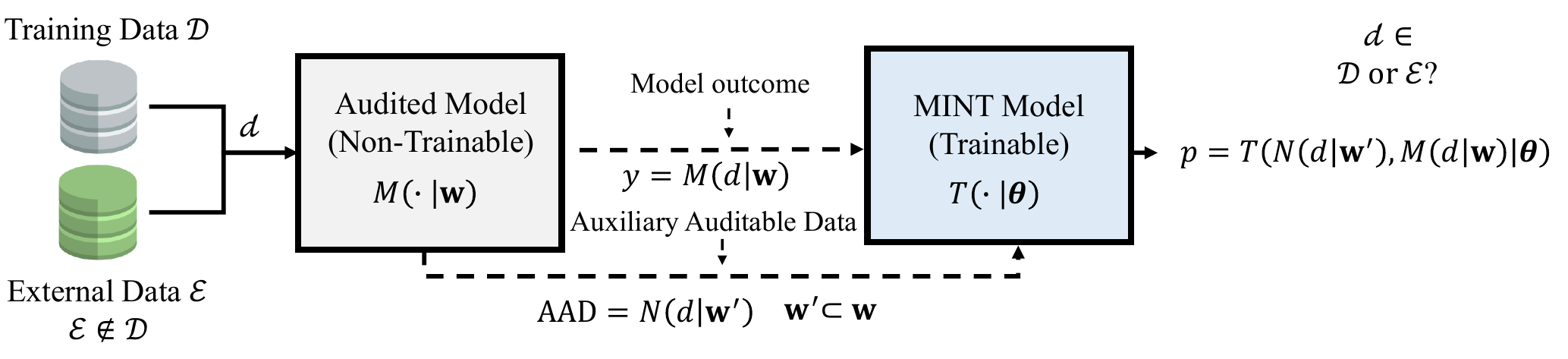} 
\caption{The Membership Inference Test (MINT) Model ($T$) is trained to predict if a specific data sample ($d$) was used during the training process of an Audited AI/ML Model ($M$), which was previously trained with a database ($\mathcal{D}$). The input of the MINT Model is AAD (e.g., activations maps for data samples $d$) and/or the model outcome obtained from $M$.}
\label{Block_Diagram_lite}
\end{figure*}

\subsection{Problem Statement and Terminology}
\label{sec:ProblState}

Let us consider a Training Dataset ($\mathcal{D}$), an External Dataset ($\mathcal{E}$) and a collection of samples ($d \in \mathcal{D} \cup \mathcal{E}$). We assume a learned model ($M$) that is trained for a specific task (e.g., text generation, Face Recognition, etc.) using the dataset $\mathcal{D}$. For any input data record ($d$), the model ($M$) generates an outcome ($y$) based on $d$ and a set of parameters ($\textbf{w}$) learned during the training process, i.e., $y=M(d|\textbf{w})$. 

We hypothesize that an authorized auditor possesses access to model $M$, enabling him the acquisition of information regarding how $M$ processes data $d$. Alternatively, the auditor may possess information detailing how $M$ has processed data $d$, even in the absence of direct model access. This information comprises the generated Model Outcome $y=M(d|\textbf{w})$ and if possible also some intermediate results (e.g., activation maps of specific layers in a neural network). These intermediate outcomes $N(d|\textbf{w}')$ provide insights about a subset of parameters $\textbf{w}' \subset \textbf{w}$. We define these intermediate outcomes as Auxiliary Auditable Data (AAD). The auditor does not need the full description of the model $M$ or the values of the complete set of parameters $\textbf{w}$ to obtain the AAD.

The aim of the proposed MINT is to determine if given data $d$ was used to train the model $M$. To this end, an authorized auditor employs a collection of AAD and/or Model Outcomes $y$ to train a MINT Model $T( \cdot| \theta)$ able to predict if a data sample $d$ belongs to the Training Data $\mathcal{D}$ or External Data $\mathcal{E}$ ($\mathcal{E} \notin \mathcal{D}$). The proposed MINT models exploit the memorization capacity of machine learning processes. The key elements of MINT are defined bellow:

\begin{itemize}
    \item Audited Model $M$: a learned model defined by an architecture and a set of parameters $\textbf{w}$.
    \item Training Data $\mathcal{D}$: collection of data used to train $M$.
    \item External Data $\mathcal{E}$: any data out of the collection $\mathcal{D}$.
    \item Model Outcome $y = M(d|\textbf{w})$: final outcome of $M$ that results from processing an input data $d$ using the set of parameters $\textbf{w}$. 
    \item Auxiliary Auditable Data $\textrm{AAD} = N(d|\textbf{w}')$: intermediate outcomes of $M$ that result from processing an input data $d$ using a subset $\textbf{w}'$ of the parameters $\textbf{w}$. The model outcome can be seen as the case where $\textbf{w}' = \textbf{w}$ in which case $N(d|\textbf{w}')=M(d|\textbf{w})$. 
    \item MINT Model $T$: a model defined by an architecture and a set of parameters $\theta$ trained using AAD $N(d|\textbf{w}')$ and/or Model Outcomes $M(d|\textbf{w})$ obtained from the two subsets of samples $\mathcal{D}$ and $\mathcal{E}$.
\end{itemize}

To provide a better comprehension, we include in Fig. \ref{Block_Diagram_lite} the complete diagram of the system's operation. 

\subsection{MINT: Application to Face Recognition Models}
\label{MINTApp2Face}

In the previous Sec. \ref{sec:ProblState}, we have introduced the general concepts of our MINT proposal, which can be applied to any type of Audited Model $M$. Next, we detail the various components of the proposal for the application of MINT to the challenging task of Face Recognition. 

The data considered in the experimental framework are face images. On one side, we have the Training Data $\mathcal{D}$, and on the other side, the External Data $\mathcal{E}$.
\begin{itemize}
    \item $\mathcal{D}$ is the Face Recognition database considered for training the Audited Model $M$.
    \item $\mathcal{E}$ consists of images from external Face Recognition databases. We could add images unrelated to faces in $\mathcal{E}$. However, we included images with similar characteristics to $\mathcal{D}$ in order to prevent the network from learning to detect database-dependent features of face images instead of focusing on the main task: database-independent detection of used or not for training. It is crucial that these images do not overlap with those in $\mathcal{D}$, i.e., $\mathcal{E} \notin \mathcal{D}$.
\end{itemize}


The audited model $M$ is a Face Recognition model. These models are trained in such a way that they learn a feature space where faces of the same person are located close together, while faces of different individuals are separated. The output of this model is not a classification vector but rather a feature embedding of the input face, within the learned feature space.


The AAD $N(d|\textbf{w}')$ consists of intermediate outputs of the network using a subset $\textbf{w}'$ of the parameters $\textbf{w}$. In this case, the Audited Model $M$ is a Face Recognition model, which primarily consists of convolutional layers. These intermediate outputs, as an application example, comprise the activation maps generated when a face image is propagated through the network during the feedforward process \cite{Serna2020Inside,Serna2022IFBiD}. The Model Outcome ($y$) is the output of the entire Face Recognition model when a face is input. As we previously explained, this will be an embedding representing that face in the multidimensional space learned by the network.

\subsection{Proposed MINT Models}
\label{MINTMod}

\begin{figure*}[!t]
\centering
\subfloat[]{\includegraphics[width=0.5\linewidth]{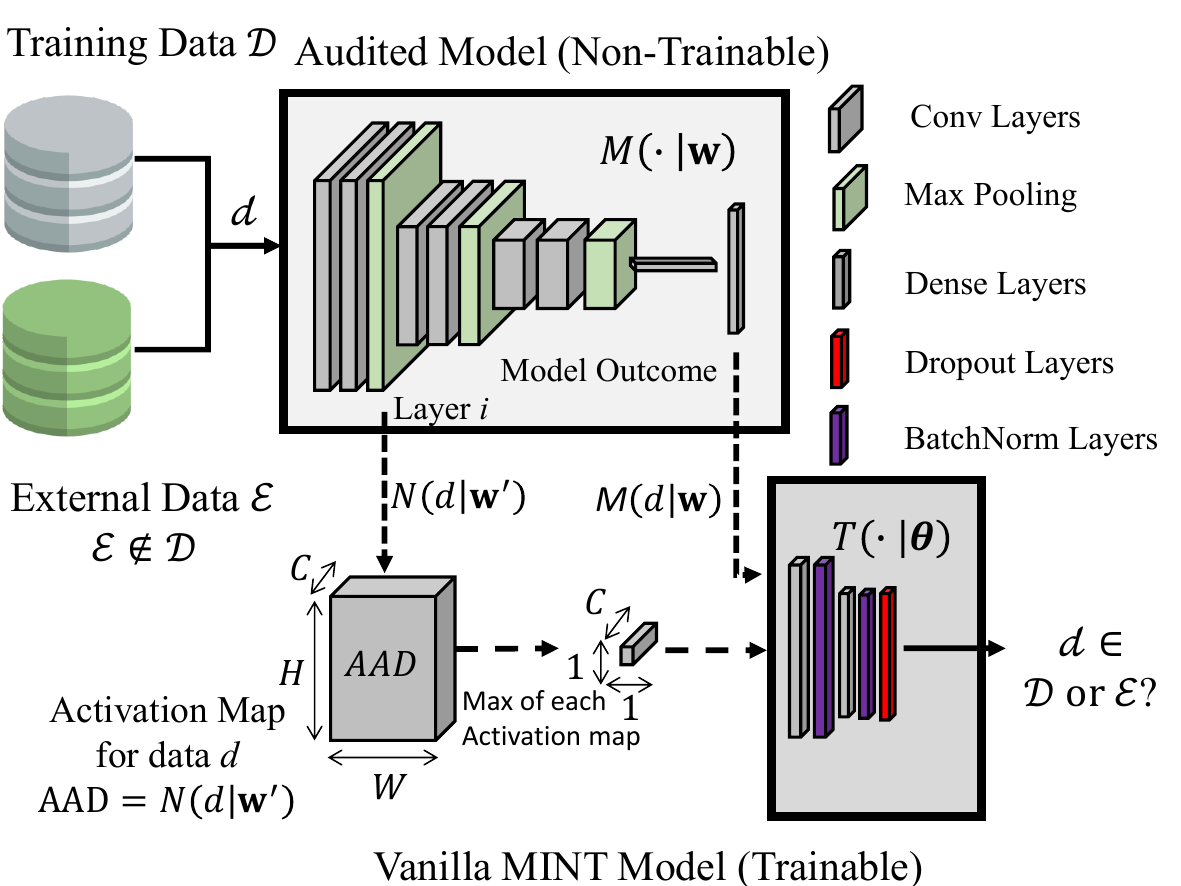}%
\label{Block_diagram_1}}
\hfil
\subfloat[]{\includegraphics[width=0.5\linewidth]{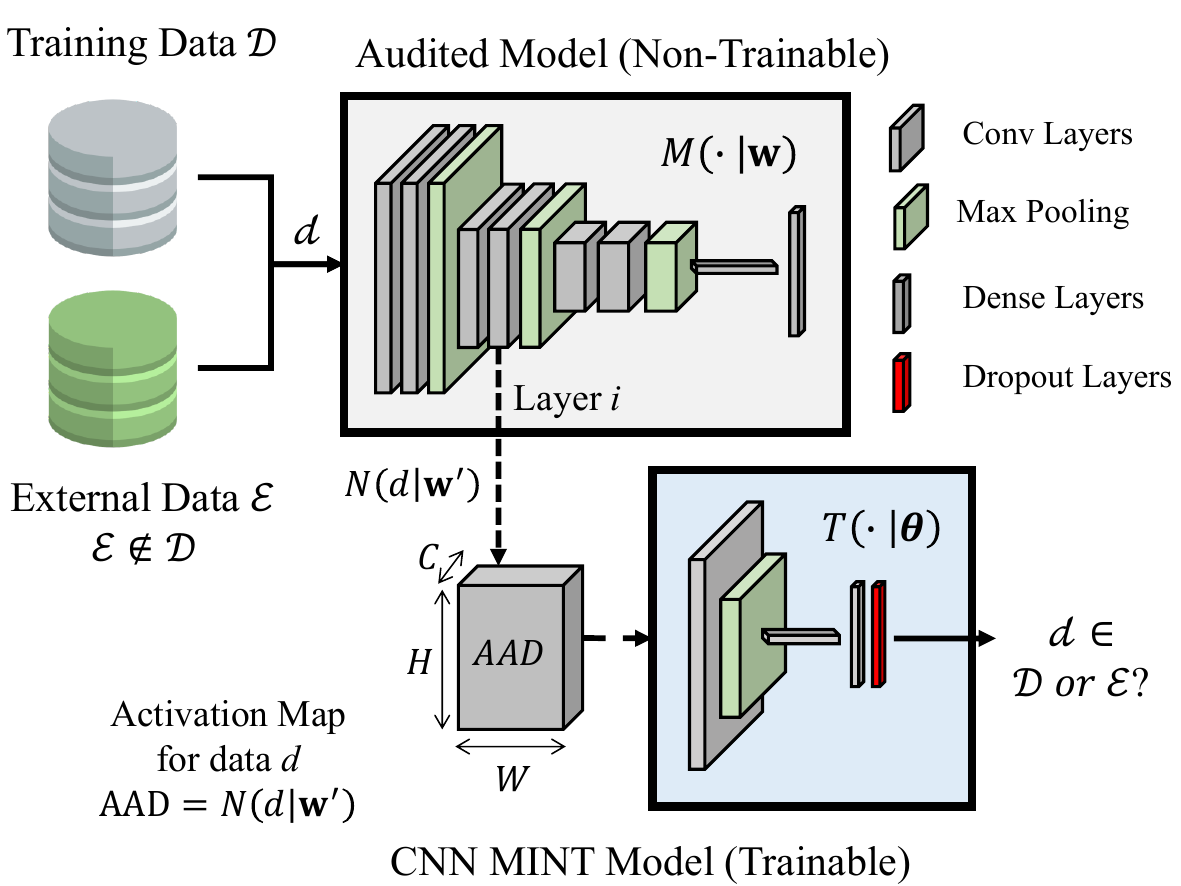}%
\label{Block_diagram_2}}
\caption{Learning framework of the Vanilla MINT Model (a) and the CNN MINT Model (b) trained with the AAD obtained from the Convolutional Layer $i$ and/or the model outcome if possible.}
\label{Block_diagram}
\end{figure*}

The architecture of the MINT Model $T$ can vary significantly based on the type of information used. First, let's focus on the format of the AAD and the Model Outcome. In the case study around face biometrics developed in the present paper, the format of activations as they pass through convolutional layers has a resolution of $H \times W$, which is influenced by padding and filter size of the convolutional layers. It also has a number of channels $C$, corresponding to the number of filters in that layer, which generally increases with deeper layers. On the other hand, the Model Outcome is simply a vector of size $L$, where $L$ is the number of dimensions/features in the learned multidimensional space. We present two MINT Models:

\begin{itemize}
    \item \textbf{Vanilla MINT Model}: We propose a MLP network to capture the training patterns within the vectorized information of the audited model ($M$). When it comes to the output embedding, it is already a vector of size $L$ that can be analyzed by the MLP. For intermediate activations to be analyzed (convolutional layers), they need to be transformed into a vector. Based on existing literature \cite{erhan2009visualizing,olah2020zoom}, data used for training maximizes the activations of intermediate neurons. Therefore, it seems reasonable to use the maximum value for each channel, obtaining $C$ values for each group of activations. Other alternatives (e.g., channel mean) were tested, but they yielded inferior results and thus were discarded. This way, we transform the activations from $H \times W \times C$ into a vector of size $C$, which can be analyzed by a MLP. If the owner of the model provides access to this information, activations and embeddings can be analyzed together or separately, depending on what yields better results. More details regarding the specific architecture of the Vanilla MINT Model are provided in Fig. \ref{Block_diagram} (a).
    \item \textbf{Convolutional Neural Network MINT Model}: We propose a CNN to capture the training patterns in the activation maps obtained from the convolutional layers of the audited model ($M$). Taking the maximum activations of each channel for MINT as done in the previous Vanilla MINT Model involves discarding a significant amount of information. Analyzing the full activations with a CNN is a more information-rich approach, which can potentially lead to better results. In Fig. \ref{Block_diagram} (b), we provide the implementation details of the proposed CNN MINT Model. Using the AAD with dimensions $H \times W \times C$ extracted from model $M$, we train a CNN with an architecture specifically designed for this input format. The utilization of the Model Outcome, which comprises vectorial information, doesn't directly fit within the MINT CNN architecture, hence it is not used as input for $T$ in this case, see Fig. \ref{Block_diagram} (b).
\end{itemize}

\section{Experimental Framework}
\label{ExperimentalFramework}

We first introduce the Face Recognition models and databases used in the present study. Then, we present the details regarding the experimental protocol.

\subsection{Face Recognition Data and Models}
\label{DataandModels}
We consider three popular Face Recognition models from the InsigthFace project \cite{insightface}. The models used ($M$ in Fig. \ref{Block_Diagram_lite}) are:
\begin{enumerate}
    \item A ResNet-100 network \cite{han2017deep}, trained on the MS1Mv3 database \cite{guo2016ms} ($\mathcal{D}$ in Fig. \ref{Block_Diagram_lite}) with ArcFace loss function \cite{deng2019arcface}. The MS1Mv3 database comprises 5.2M images from 91K identities.
    \item A ResNet-100 network, trained on the Glint360K database \cite{an2021partial} ($\mathcal{D}$) with CosFace loss function \cite{wang2018cosface}. The Glint360K database comprises 17M images from 360K identities.
    \item A partial fully-connected ResNet-100 network \cite{an2022pfc}, trained on the Glint360K database ($\mathcal{D}$) with CosFace loss function.
\end{enumerate}

As external data ($\mathcal{E}$) to train and test the MINT model $T$ we use the following datasets: The IJB-C \cite{IJB} (3.5K identities and 138K face images), FDDB \cite{jain2010fddb} (5.2K face images), GANDiffFace \cite{melzi2023gandiffface, melzi2024frcsyn} (10K identities and 500K face images), and Adience \cite{Adience} (2.2K identities 26.5K face images). As mentioned in Sec. \ref{MINTApp2Face}, it is crucial to avoid overlap between $\mathcal{E}$ and $\mathcal{D}$, as this would lead to incorrect labels, compromising the training of MINT Models $T$. As noted by Schlett \textit{et al.} \cite{icpram24}, there is a significant number of duplicates across commonly used datasets. Although the data in $\mathcal{E}$ is sourced from evaluation datasets that strive to prevent such overlap, the necessary tests have been conducted to ensure that this issue does not occur.

\subsection{Experimental Protocol}
\label{exp_prot}

Below, we elaborate on the diverse factors considered in this study to establish an equitable experimental protocol:

\begin{itemize}
    \item No identity overlap when dividing the data between Train and Evaluation, ensuring that performance is not influenced by identity recognition rather than the primary task of recognizing training and non-training data.
    \item $4$ different databases as External Data $\mathcal{E}$. Three of them (IJB-C, FDDB, GANDiffFace) are used to extract samples for training the MINT Model, while the fourth one (Adience) is used for evaluation. We will refer to this choice of databases as the ``Baseline Case''. Having either IJB-C or GANDiffFace, or both, in the training set is necessary to achieve balanced training (enough $\mathcal{E}$ samples). If we include only GANDiffFace from these two databases in the training set, the network might learn to recognize specific patterns from synthetic images instead of focusing on the MINT task. Therefore, we always include IJB-C in the training set. However, the remaining three databases are interchangeable, allowing evaluation with GANDiffFace or FDDB and utilization of Adience in training. We refer to these two cases as ``Case A'' and ``Case B'', respectively. We present some results interchanging this evaluation database in our experiments. The Baseline case provides the most robust evaluation, as Case A involves evaluation with synthetic images, which may raise doubts due to the nature of these images, and Case B involves evaluation with only 5.2K images. Therefore, the majority of results will be presented under the ``Baseline Case''.
    \item In the evaluation, there is an equal number of samples for each class (the same from $\mathcal{D}$ and $\mathcal{E}$) to ensure balanced evaluation results. As the Adience database $E$ used for evaluation in ``Baseline Case'' has 26.5K images, we took the same number of images from $D$, leaving the rest of the $D$ database available for training. 
    
\end{itemize}

\subsubsection{Data Pre-processing}

We first apply the same face detector to the images of all databases. The face detector used can be found in the InsightFace project \cite{insightface}, and it is the SCRFD Detector \cite{guo2021sample}. This face detector is the same one used for the training of the Face Recognition models. With this face detector, we extract faces from the images, crop, align them, and then resize them to $112 \times 112$. This ensures that all the images consist of a face in the center of the image, with the same orientation and size. This way, we establish similar conditions for all the images thereby promoting focus on the MINT task.

\subsubsection{Auxiliary Auditable Data and Model Outcome}
\label{AADYMO}
Once the data is preprocessed, we proceed with the feedforward pass through the Face Recognition network. We feed the face images from datasets $\mathcal{D}$ and $\mathcal{E}$ into the network $M$ and extract the AAD and the Model Outcome. For the Model Outcome, we simply retrieve the output feature embedding. The AAD is derived from intermediate outputs of the model. Therefore, we must determine the depth within the model to obtain this AAD. The three Face Recognition models are based on the ResNet100 architecture as proposed by Han \textit{et al.}  \cite{han2017deep}. This architecture comprises four primary layers, each consisting of the designated Building Blocks. In this study, we collect the output of the final convolutional layer in the last Building Block of each of the four primary layers, resulting in four distinct AAD groups.

\subsubsection{MINT Models Training}
\label{MINTMODEL}

With the previously obtained AAD and Model Outcome, we propose two MINT models. Depending on how we handle the acquired information, we propose two architectures (Sec. \ref{MINTMod}): a Vanilla MLP model and a CNN model. The CNN utilizes the activations of the convolutional blocks of $M$ as-is, while the Vanilla model, as previously explained, applies max pooling per channel to convert the activations of the convolutional blocks into a vector that can be processed by the MLP.

The specific architecture of the Vanilla model consists of a dense layer with 128 neurons, ReLU activation, and an L1 regularizer with a value of 0.1. This is followed by a batch normalization layer, a dropout layer with a value of 0.5, and finally, the output layer with 1 neuron and sigmoid activation. The training algorithm consists of Adam optimizer with default settings and $lr=0.001$, training for $20$ epochs with batches of 128 samples, each with an equal number of samples per class. Binary cross-entropy is employed as the loss function.

The CNN architecture consists of a convolutional layer with 64 filters, each of dimension $C$ with a kernel = 5, and ReLU activation. The architecture is completed by a max-pooling layer and, finally, a fully connected layer with $C$ neurons, ReLU activation, and a dropout of 0.5 followed by the output layer with 1 neuron and sigmoid activation. The training algorithm consists again of Adam optimizer with default settings and $lr=0.001$, training for $30$ epochs with batches of 128 samples, each with an equal number of samples per class. Binary cross-entropy as the loss function.

Both architectures are chosen after extensive experimentation. It was found that the architectures did not need to be overly complex, as performance did not improve significantly with more complex architectures. In the experimental section (Section \ref{Results}), an experiment has been included to demonstrate the model's performance versus the network complexity.

\section{Experiments and Results}
\label{Evaluation}

This section presents the results achieved by our proposed MINT approach. As it involves binary classification between $\mathcal{D}$ and $\mathcal{E}$, and considers a balanced evaluation, the baseline accuracy (i.e., random guess) stands at 50\%. Additionally, we recall that all results are derived from samples in $\mathcal{E}$ that do not belong to the $\mathcal{E}$ databases used in training the MINT Model, as elaborated in Sec. \ref{exp_prot}.

The nomenclature used in this section is as follows:

\begin{itemize}
    \item \textit{FR Model ($M$ in Fig. \ref{Block_Diagram_lite}):} These are the Face Recognition models presented in Sec. \ref{DataandModels}. The number corresponds to the one specified in that section.
    \item \textit{Model Outcome:} Refers to the Output Embedding of the Face Recognition model in Sec. \ref{MINTApp2Face}.
    \item \textit{Conv Layer X:} Represents the AAD obtained from each of the four primary layers, as explained in Sec. \ref{AADYMO}
    \item \textit{Combination:} Indicates the results using the AAD obtained from the four primary layers. The combination, including the Model Outcome as well, yields exactly the same results.
    \item \textit{Training Data:} Represents three possible scenarios based on the amount of data available for training our MINT Model $T$. The high scenario ($100$K) assumes a significant amount of data, with $50$K data $\mathcal{D}$ and $50$K external data $\mathcal{E}$. In the medium scenario ($50$K), there are $25$K data samples each for $\mathcal{D}$ and $\mathcal{E}$. The low scenario ($1$K) considers only $500$ data samples for both $\mathcal{D}$ and $\mathcal{E}$. To put these numbers in context, $50$K data samples from $\mathcal{D}$ represent only $1\%$ of the MS1Mv3 database used for training and $0.3\%$ of Glint360k.
\end{itemize}

\subsection{Results}
\label{Results}

Table \ref{Table:AccForLayerNN} shows the classification accuracy results for the $3$ Face Recognition models using different information to train the Vanilla MINT Model $T$ (AAD and Model Outcomes). All of this is done under a scenario of high data available for training. In these results, we can observe a superiority in accuracy when combining all the AAD information (Combination). In terms of the AAD it is important to highlight that in FR Model 1, Conv Layer \#2 yields the best results after Conv Layer \#1, while in FR Model 2 and 3, Conv Layer \#4 follows with better accuracy. This might be attributed to the fact that ResNet architectures are based on ``Skip Connections'', allowing information to traverse through the network from first layers to the last ones \cite{he2016deep}. 

\begin{table}
\begin{center}
\caption{Classification accuracy for various AAD and Model Outcomes using the Vanilla MINT Model. MINT model trained with $100K$ samples ($1\%$ of the total Face Recognition Model training set for FR Model 1 and $0.3\%$ for FR Models 2 and 3).}
\label{Table:AccForLayerNN}
\begin{tabular}{@{}lccc@{}}
\hline
Auditable Data & \multicolumn{1}{l}{FR MODEL 1} & FR MODEL 2 & FR MODEL 3 \\ 
\hline
Conv Layer \#1 & 0.77 & 0.80 & 0.78 \\
Conv Layer \#2 & 0.74 & 0.68 & 0.68 \\
Conv Layer \#3 & 0.68 & 0.59 & 0.62 \\
Conv Layer \#4 & 0.69 & 0.76 & 0.75 \\ 
Model Outcome & 0.75 & 0.78 & 0.76 \\
\textbf{Combination} & \textbf{0.84} & \textbf{0.84} & \textbf{0.81} \\ 
\hline
\end{tabular}
\end{center}
\end{table}

The results in this section pertain to the Baseline Case presented in Sec. \ref{exp_prot}, where the databases IJB-C, FDDB, and GANDiffFace are used in training the MINT Models, while Adience is used for evaluation. This is the case that provides a more robust evaluation (Sec. \ref{exp_prot}), and therefore the one mostly used throughout the experimental section. However, to demonstrate the robustness of the results, Table \ref{Table:AccForLayerNNCases} showcases the same results as Table \ref{Table:AccForLayerNN} with exchanged training and evaluation databases for the other two scenarios (see Sec. \ref{exp_prot}):
\begin{itemize}
    \item \textbf{Case A}: IJB-C, FDDB, and Adience are used for training, with GANDiffFace for evaluation.
    \item \textbf{Case B}: IJB-C, GANDiffFace, and Adience are used for training, with FDDB for evaluation.
\end{itemize} 
As shown in Table \ref{Table:AccForLayerNNCases}, the outcomes are slightly superior for Case A and slightly inferior for Case B compared to the Baseline Case, yet the the conclusions remain unchanged (results are presented as Case A / Case B).

\begin{table}
\begin{center}
\caption{Classification accuracy for various AAD and Model Outcomes using the Vanilla MINT Model. MINT model trained with $100K$ samples ($1\%$ of the total Face Recognition Model training set for FR Model 1 and $0.3\%$ for FR Models 2 and 3). We display the results for Cases A and B (Case A/Case B in Table).}
\label{Table:AccForLayerNNCases}
\begin{tabular}{@{}lccc@{}}
\hline
Auditable Data & \multicolumn{1}{l}{FR MODEL 1} & FR MODEL 2 & FR MODEL 3 \\ 
\hline
Conv Layer \#1 & 0.84/0.78 & 0.83/0.75 & 0.86/0.75 \\
Conv Layer \#2 & 0.68/0.68 & 0.65/0.67 & 0.63/0.66 \\
Conv Layer \#3 & 0.54/0.53 & 0.51/0.58 & 0.53/0.59 \\
Conv Layer \#4 & 0.53/0.55 & 0.77/0.75 & 0.71/0.72 \\ 
Model Outcome & 0.73/0.64 & 0.86/0.79 & 0.84/0.74 \\
\textbf{Combination} & \textbf{0.88/0.78} & \textbf{0.87/0.82} & \textbf{0.90/0.80} \\ 
\hline
\end{tabular}
\end{center}
\end{table}

In Table \ref{Table:AccForScenarioNN}, we can observe the classification accuracy when we vary the data available to train the Vanilla MINT model for the best obtained Auditable Data (combination of all convolutional layers). Naturally, accuracy decreases when we reduce the number of training samples, demonstrating that the patterns to be learned are not trivial. A noteworthy result is that in the $50K$ scenario, accuracy decreases very slightly with respect to $100K$. This result suggests that an intermediate number of training samples may be enough to achieve acceptable results. Another interesting result is what occurs in the $1K$ scenario. For FR Model 1, performance decreases by $18\%$. However, for FR Model 2 and FR Model 3, it decreases by only $8\%$ and $6\%$, respectively. From these results, we can infer that depending on the model $M$ (its training, architecture, loss function, etc.), it may be possible to develop our MINT Model $T$ with more or less training data.

\begin{table}
\begin{center}
\caption{Classification accuracy for the Vanilla MINT Model (Combination as Auditable Data) in the various data provided scenarios discussed for the three Face Recognition (FR) models.}
\label{Table:AccForScenarioNN}
\begin{tabular}{@{}lccc@{}}
\hline
Training Data & FR Model 1 & FR Model 2 & FR Model 3 \\
\hline
$100$K & 0.84 & 0.84 & 0.81 \\
$50$K & 0.84 & 0.82 & 0.79 \\
$1$K & 0.66 & 0.76 & 0.75 \\ 
\hline
\end{tabular}
\end{center}
\end{table}

In Table \ref{Table:AccForLayerCNN}, we can observe the different classification accuracy results for the $3$ Face Recognition Models using the different information available to train the CNN MINT Model $T$ under the high data scenario. Here, the CNN is trained using the full activation blocks of each convolutional layer. Once again, the best performance is achieved with the Combination of all Conv Layers. Similarly, as observed previously, the next best results are achieved with Conv Layer \#1 (i.e., the layer closest to the image domain), progressively decreasing as the layer gets closer to the end of the network (i.e., the layer closest to the output domain). The Model Outcome is not used, as explained in Sec. \ref{MINTMod}, because, as we discussed there, it is vectorial information that doesn't fit in a CNN architecture in a straightforward way. For these CNN MINT Models, training with the combination of all the AADs significantly increases the computational complexity (approximately fourfold), while yielding only a modest improvement of around 1\% in accuracy. Consequently, for the remaining experiments, we proceed with the CNN MINT Model trained using Conv Layer 1 as the AAD.


  

\begin{table}
\begin{center}
\caption{Classification accuracy for various AAD using the CNN MINT model. MINT model trained with $100$K samples ($1\%$ of the total Face Recognition Model training set for FR Model 1 and $0.3\%$ for FR Models 2 and 3).}
\label{Table:AccForLayerCNN}
\begin{tabular}{@{}lccc@{}}
\hline
Auditable Data & FR Model 1 & FR Model 2 & FR Model 3 \\ 
\hline
Conv Layer \#1 & 0.86 & 0.89 & 0.90 \\
Conv Layer \#2 & 0.85 & 0.86 &  0.86\\
Conv Layer \#3 & 0.83 & 0.75 &  0.76 \\
Conv Layer \#4 & 0.73 & 0.74 &  0.73\\ 
\textbf{Combination} & \textbf{0.88} & \textbf{0.90} & \textbf{0.90}\\
\hline
\end{tabular}
\end{center}
\end{table}

In Table \ref{Table:AccForScenarioCNN}, we observe the impact of varying the number of available data used to train the CNN MINT Model $T$ for the considered Auditable Data (Conv Layer \#1). We see that for FR Model 1, a scenario with limited data for training the MINT Model negatively affect our results. However, for the other two FR Models, it has almost no effect on accuracy, achieving consistently high accuracy despite decreasing significantly the amount of training data. This indicates that for this MINT architecture depending on the model being audited, few samples ($500$ from $\mathcal{D}$) are enough to achieve good results. This result is very interesting as it implies that the auditor would require little cooperation from the developer. A similar situation occurs in Table \ref{Table:AccForScenarioNN} with the Vanilla MINT Model. The reason behind this is that, in Model 1, the activation patterns indicating whether a sample was used for training or not are more complex than in Models 2 and 3. As a result, the MINT Model requires more training data compared to the other two models to perform effectively.

\begin{table}
\begin{center}
\caption{Classification accuracy for the CNN MINT Model (Conv Layer \#1 as Auditable Data) in the various data provided scenarios discussed for the three Face Recognition (FR) models.}
\label{Table:AccForScenarioCNN}
\begin{tabular}{@{}lccc@{}}
\hline
Training Data & FR Model 1 & FR Model 2 & FR Model 3 \\ 
\hline
$100$K & 0.86 & 0.89 & 0.90 \\
$50$K & 0.85 & 0.89 & 0.87 \\
$1$K & 0.73 & 0.87 & 0.87 \\ 
\hline
\end{tabular}
\end{center}
\end{table}

To conclude we compare our proposed MINT model to the state of the art, in Table \ref{Table:ComparisonSOTA}. As mentioned earlier, to the best of our knowledge, there are no studies addressing the problem of detecting whether specific data was used for training state-of-the-art Face Recognition models. Therefore, making direct comparisons with other studies is not possible. Nevertheless, most studies around MIAs often use the classification output vector of the network \cite{shokri2017membership, yeom2018privacy, song2021systematic, salem2018ml}, or the output from the layers just before this output \cite{nasr2019comprehensive, rezaei2021difficulty}. These approaches are somewhat analogous to our Model Outcome (the Model Outcome of a Face Recognition Model is similar to the layers just before the output vector in a classification model), and thus, we use them as a reference for interpreting the state of the art, despite the environmental differences between the works around MIAs and our proposed MINT.

\begin{table}
\begin{center}
\caption{Classification accuracy comparison between MIA and the two MINT approaches proposed in this work. $^{*}$We have adapted the MIA approach used in \cite{rezaei2021difficulty} using the Face Recognition Model outcome. Note that we used the original model instead of shadow models. $^{**}$We used the Vanilla MINT model based on the combination of all convolutional layers. $^{***}$We used the CNN MINT Model trained with the Conv Layer \#1.}
\label{Table:ComparisonSOTA}
\begin{tabular}{@{}lccc@{}}
\hline
Method & FR Model 1 & FR Model 2 & FR Model 3 \\ \hline
MIA \cite{rezaei2021difficulty}$^*$  & 0.75 & 0.78 & 0.76 \\
MINT Vanilla$^{**}$ & 0.84 & 0.84 & 0.81 \\
\textbf{MINT CNN}$^{***}$ & \textbf{0.86} & \textbf{0.89} & \textbf{0.90} \\ \hline
\end{tabular}%
\end{center}
\end{table}

\begin{figure*}
\centering
\subfloat[]{\includegraphics[width=0.33\linewidth]{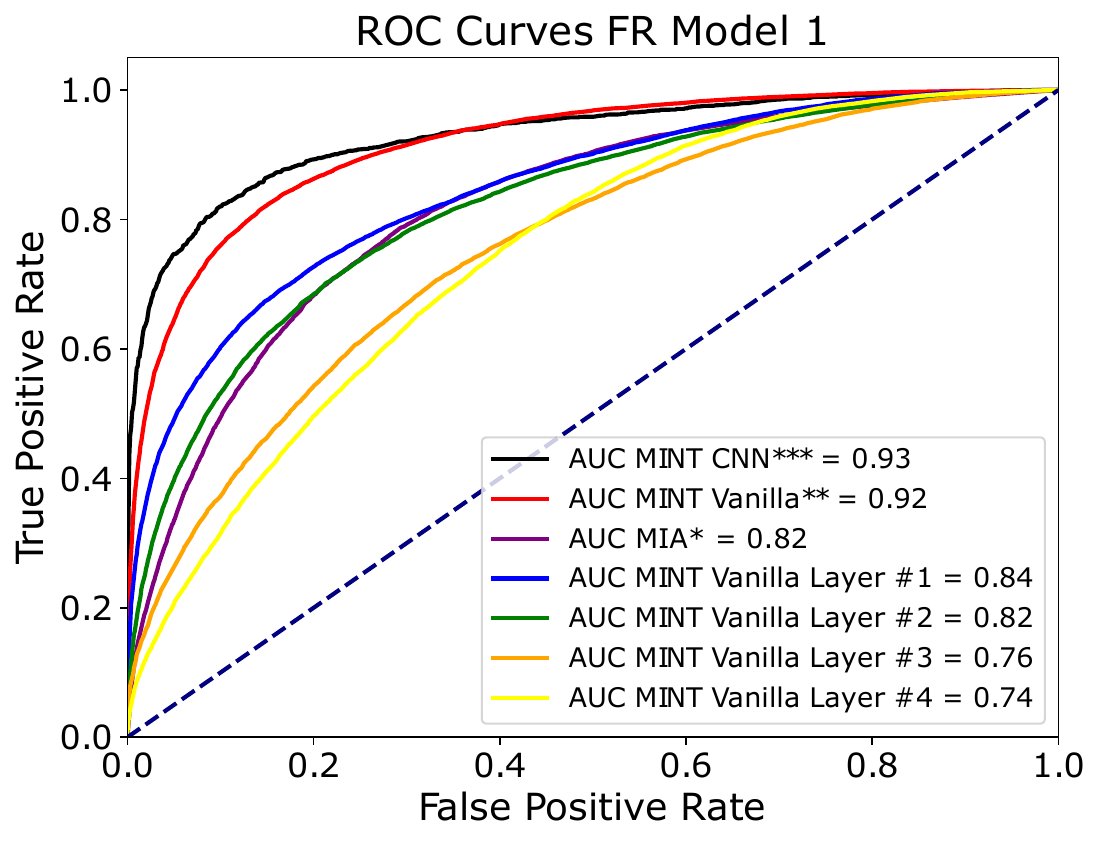}%
\label{fig:rocfinal_1}}
\hfil
\subfloat[]{\includegraphics[width=0.33\linewidth]{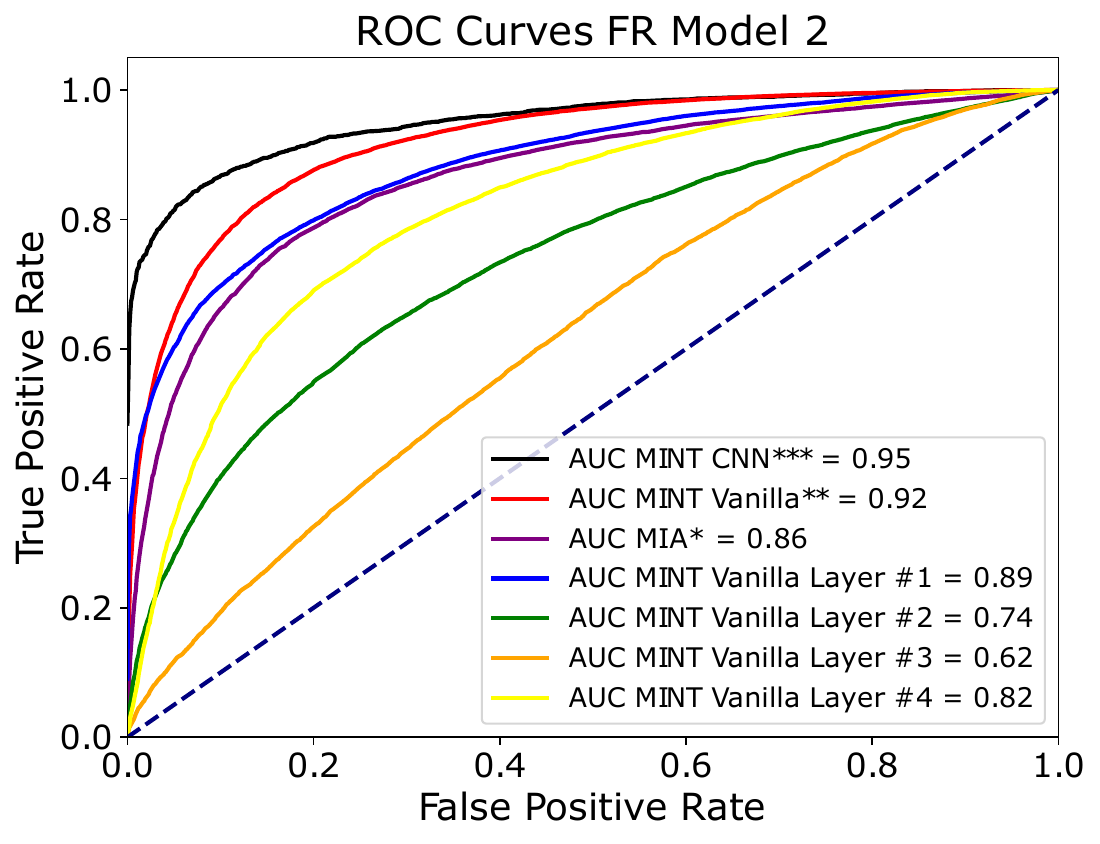}%
\label{fig:rocfinal_2}}
\hfil
\subfloat[]{\includegraphics[width=0.33\linewidth]{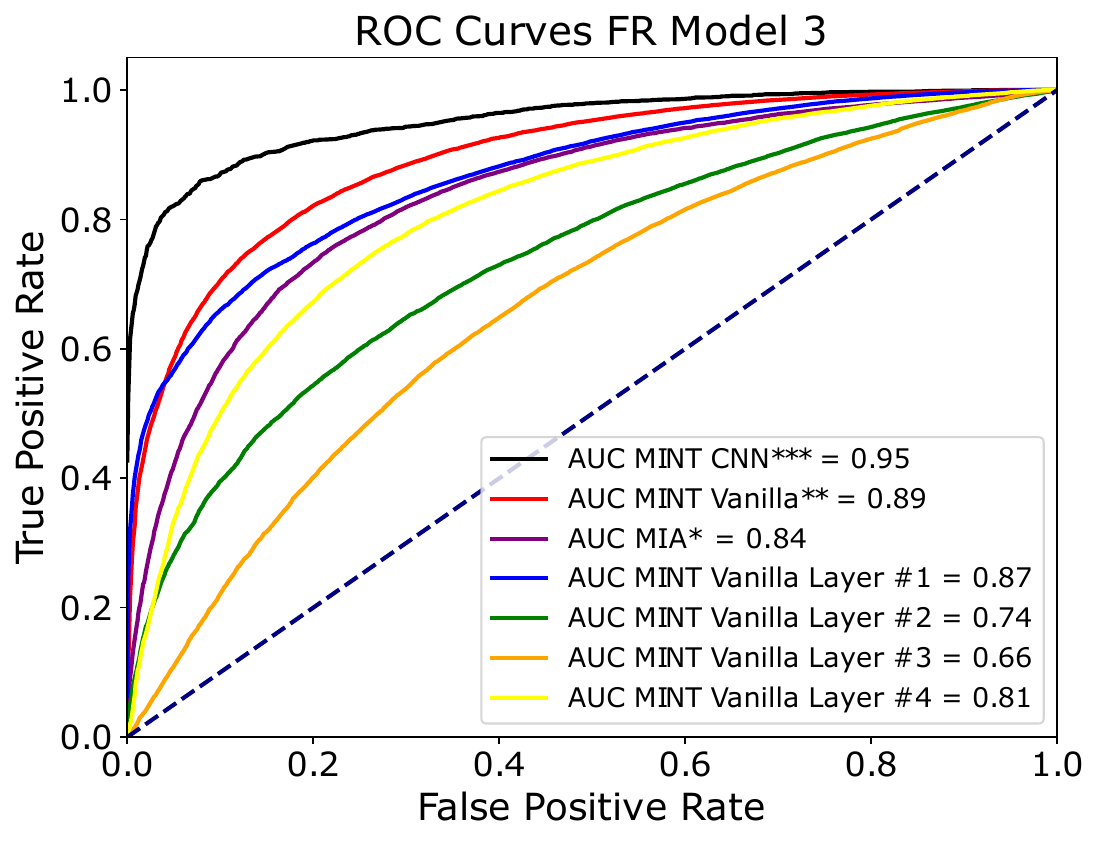}%
\label{fig:rocfinal_3}}
\caption{ROC curves obtained of the different MINT approaches for the FR Model 1 (left), FR Model 2 (center), and FR Model 3 (right). $^{*}$We have adapted the MIA approach used in \cite{rezaei2021difficulty} using the Face Recognition Model outcome. Note that we used the original model instead of shadow models. $^{**}$We used the Vanilla MINT model based on the combination of all convolutional layers. $^{***}$We used the CNN MINT Model trained with the Conv Layer \#1.}
\label{fig:rocfinal}
\end{figure*}

Subsequently, we compare in Table \ref{Table:ComparisonSOTA} the best published MIA results with the best results achieved in our present study. As observed, the methods proposed in the present study significantly outperform what could be considered the state of the art in Face Recognition MIAs, surpassing it for all three FR Models by more than $10\%$. Finally, Fig. \ref{fig:rocfinal} provides a comparison of the ROC curves for the three methods outlined in Table \ref{Table:ComparisonSOTA}, along with the results for the Vanilla network with information from individual Conv Layers. An important leap is evident between MIA (purple curve) and our best result using the CNN MINT Model (black curve), which we have also observed in Table \ref{Table:ComparisonSOTA}. One of the main criticisms discussed in \cite{rezaei2021difficulty} and \cite{carlini2022membership} is that the evaluation protocol and metrics commonly used in MIA approaches are not adequate, and the results presented in previous studies suffer from a high False Positive Rate. In Fig. \ref{fig:rocfinal}, we can see that this is not the case in our study.

As explained in Sec. \ref{MINTMODEL}, an in-depth heuristic analysis (not fully reported here) has been conducted varying the complexity of the presented MINT Model architectures. From that study we can conclude that increasing the complexity of the architecture does not easily improve the performance. In our experiments, more complex architectures led to overfitting and a loss of performance in evaluation. To demonstrate this, Table \ref{Table:ComparisonParameters} presents results by varying the number of parameters of the MINT CNN architecture. We provide results with the original model as explained in Sec. \ref{MINTMODEL}, another architecture with approximately three times the number of parameters, and the last one with ten times the number of parameters. We also include an experiment with a model three times simpler to demonstrate that the CNN requires enough complexity. Finally, we also present results using two additional architectures widely employed in the state of the art in computer vision. Specifically, the results obtained using a ResNet50 \cite{he2016deep} and a ViT \cite{alexey2020image} as MINT Models.

\begin{table}
\begin{center}
\caption{Classification accuracy comparison while varying the complexity of the MINT CNN network used. $^{*}$ CNN MINT architecture with $\div3$ parameters. $^{**}$ Original MINT CNN architecture presented in \ref{MINTMODEL}. $^{***}$ CNN MINT architecture with $\times3$ parameters. $^{****}$ CNN MINT architecture with $\times10$ parameters. $^{*****}$ ResNet50 \cite{he2016deep} as CNN MINT architecture. $^{******}$ ViT \cite{alexey2020image} as CNN MINT architecture}
\label{Table:ComparisonParameters}
\begin{tabularx}{\linewidth}{@{}lXXXX@{}}
\hline
Method & \#Param & FR 1 & FR 2 & FR 3 \\ \hline
MINT CNN$^{*}$& $580$ & 0.79 & 0.81 & 0.82 \\ \hline
MINT CNN$^{**}$& $1.7$K & 0.86 & 0.89 & 0.90 \\ \hline
MINT CNN$^{***}$ & $5$K & 0.86 & 0.88 & 0.90 \\ \hline
MINT CNN$^{****}$ & $18$K & 0.85 & 0.88 & 0.88 \\ \hline
MINT ResNet50 & $25.6$M & 0.86 & 0.89 & 0.89 \\ \hline
MINT ViT & $86$M & 0.86 & 0.89 & 0.90 \\ \hline
\end{tabularx}
\end{center}
\end{table}


\section{Why Face Recognition? Exploring its Relevance and Broader Applications}
\label{sec:whyface}

In this initial work, the decision was made to focus on the field of computer vision (CV) due to the wide availability of datasets and models and the significant relevance of CV within AI. However, applying this technology to fields like NLP is also highly interesting, as NLP often requires an immense amount of data, which can sometimes be subject to copyright protection.

Within computer vision, face recognition (FR) is one of the areas where leveraging massive amounts of data could be particularly beneficial for improving systems \cite{Wang2023masks, Masi2016fr}. This need may tempt developers to use data without proper authorization. FR is also highly sensitive to variability and changes, often necessitating model retraining to adapt to new situations—e.g., the use of masks during the COVID-19 pandemic, which prompted adjustments in face recognition systems \cite{Wang2023masks}. The need for rapid updates can further encourage the use of data without adequate consent. Additionally, facial images in CV are highly sensitive, often met with public opposition \cite{lynch2020face,Yueming20233D,Melzi2023WACV,Serna2022Sensitive} and protected by various legal frameworks \cite{GDPR,CCPA, BIPA}. These factors collectively motivated us to introduce MINT in the FR domain.

Nevertheless, within CV, we have also applied MINT to image classification tasks. Although image classification might be less visually appealing than face recognition and its utility may seem less obvious, it remains critical. Many images, not limited to faces, may belong to individuals and be protected under copyright—e.g., landscape photos taken by a photographer—making their unauthorized use a potential violation of legislation and individual rights. Moreover, MINT is a valuable tool for international competitions, where the goal is often to improve model classification performance, and it is crucial that MINT performs effectively in these tasks to identify whether participants have trained models with unauthorized data. Experiments in the domain of image classification can be found in Appendix A.

\section{Real-World Deployment of MINT: Challenges and Legal Implications}
\label{sec:realworld}

\subsection{Obfuscation Methods}

MINT is a tool designed to identify data that has been used to train a model, aiming to protect citizen rights and promote safer and more trustworthy AI. It seeks to prevent developers from leveraging data without the required permissions. However, this raises an important question: Can developers use unauthorized data and deliberately conceal its use from MINT Models?

In this work, we demonstrate that detection of training data is feasible even with state-of-the-art models. Nevertheless, developers aware of MINT's capabilities could design new training techniques to obscure the presence of unauthorized data, which would necessitate further advancements in MINT methods. This interplay opens a future research avenue, requiring MINT approaches to evolve alongside emerging concealment techniques.


\subsection{Deployment and Legal Implications}

The deployment process of MINT could work as follows: a developer releases a new model $M$ (e.g. a FR Model), and an auditing entity (e.g., the EU or a private organization) trains the MINT model $T$ and makes it publicly available. This auditing entity could verify whether certain popular databases or internet-scraped data were used in training the model $M$. Simultaneously, users can test the model to determine if their images, such as those shared on social media, were used without consent and report such findings to the auditing entity. If it is detected that a developer trained a model using data without proper authorization, severe legal, ethical, and reputational consequences could follow.

From a legal perspective, particularly regarding biometric data like facial images, many countries enforce regulations that protect these sensitive data types, such as the GDPR \cite{GDPR}, CCPA \cite{CCPA}, and BIPA \cite{BIPA}. These laws can impose substantial financial penalties in addition to potential class-action lawsuits. Furthermore, if the data were sourced from databases or platforms, this could violate their terms of service and copyright laws. Developers may also face individual or collective lawsuits from affected individuals. 

On the other hand, developers would also face ethical and reputational consequences, leading to a loss of trust from users in both the developers and their associated companies, as well as generating negative publicity.

Furthermore, legal penalties could include model withdrawal or access restrictions, while reputational harm might make it difficult for developers to launch future models that users trust and adopt.

\section{Discussion}
\label{Discussion}

It is important to contextualize the results achieved in the present study. While in Sec. \ref{Results} we provide a comparison with the state of the art in MIAs, this is for a simple interpretation as there is no specific work that aims at detecting if given data was used for training on state-of-the-art Face Recognition Models. In the state of the art of MIAs, results are available for classification tasks with small databases such as CIFAR-10 \cite{krizhevsky2009learning} where accuracy reaches up to 0.631 \cite{watson2021importance} or Yale Face \cite{598228} where accuracy is 0.558 \cite{song2019privacy}. The complexity of state-of-the-art Face Recognition models and databases is much higher, and with more relevant real-world application. Furthermore, the results achieved with the proposed MINT approach reach up to $90\%$.

On the other hand, it is important to emphasize the context of our study and the results achieved. As explained in Sec. \ref{sec:MINT}, existing MIAs and our proposed MINT operate under different environmental conditions. These differences arise from the distinct application scenarios of both technologies. While MIA is designed as an attack, MINT is conceived as an auditing tool motivated on the recent legislation and aims for user transparency. As a result, MIAs require the creation of shadow models to simulate the behavior of the original model to attack, whereas in MINT we have direct access to the original model. This leads to divergent outcomes and the potential utilization of information in MINT that is not particularly useful in MIA. For instance, the use of intermediate activations seems not to yield substantial improvements in MIAs \cite{rezaei2021difficulty,cretu2023re,nasr2019comprehensive}. Nevertheless, as demonstrated in our study, appropriately leveraging this information significantly enhances performance in MINT. This implies that ineffective methods in MIA do not necessarily perform poorly in MINT, suggesting opportunities for technology-specific advancements. As an example, exploring the use of gradients could be a promising avenue for future research in MINT.

\section{Conclusions}
\label{Conclusions}

This paper has presented the Membership Inference Test (MINT), a novel approach to empirically assess whether specific data was used in the training of Artificial Intelligence (AI) models from an auditor's standpoint. Our research introduces two innovative architectures for modeling activation patterns: one based on Multilayer Perceptron (MLP) networks and the other on Convolutional Neural Networks (CNNs). These models have been rigorously evaluated in the challenging context of Face Recognition, using three state-of-the-art Face Recognition models and six publicly available databases comprising over 22 million face images.

Our findings demonstrate that our proposed MINT approach can achieve up to $90\%$ accuracy in identifying whether a particular AI model has been trained with specific data. This is a significant advancement, especially considering the increasing demands for transparency and accountability in the use of AI technologies. 


This work opens numerous avenues for future research. A direct path is to enhance the presented work by incorporating additional information like the gradients of the data. The study demonstrates the effectiveness of the MINT Model even with only $1$K training samples; a subsequent step would be to explore unsupervised training and identify patterns in the activations in an unsupervised manner, eliminating the need for developer involvement. Another future direction could involve training the MINT Model alongside the Audited Model with a joint loss function to optimize performance for both tasks simultaneously. This would necessitate active collaboration from the model developer, leading to what we might call Active Membership Inference Testing, as opposed to the Passive Membership Inference Testing presented in this work. Additional lines of research include applying this technology to other applications, such as image generation models \cite{Joao2022GAN,HASSANPOUR2024110442}, or to different types of data, such as text \cite{PENA2024102398}, and studying MINT on widely available LLMs \cite{ivan24gpt}.

\section*{Acknowledgement}
\label{Acknowledgement}

This study has been supported by projects BBforTAI (PID2021-127641OB-I00 MICINN/FEDER), HumanCAIC (TED2021-131787B-I00 MICINN), M2RAI (PID2024160053OB-I00 MICIU/FEDER) and Cátedra ENIA UAM-VERIDAS en IA Responsable (NextGenerationEU PRTR TSI-100927-2023-2). The work of D. deAlcala is supported by a FPU Fellowship (FPU21/05785) from the Spanish MIU. A. Morales is supported by the Madrid Government (Comunidad de Madrid-Spain) under the Multiannual Agreement with Universidad Autónoma de Madrid in the line of Excellence for the University Teaching Staff in the context of the V PRICIT (Regional Programme of Research and Technological Innovation). The work has been conducted within the ELLIS Unit Madrid.

\bibliographystyle{elsarticle-num}
\bibliography{main}

\EOD

\begin{IEEEbiography}[{\includegraphics[width=1in,height=1.25in,clip,keepaspectratio]{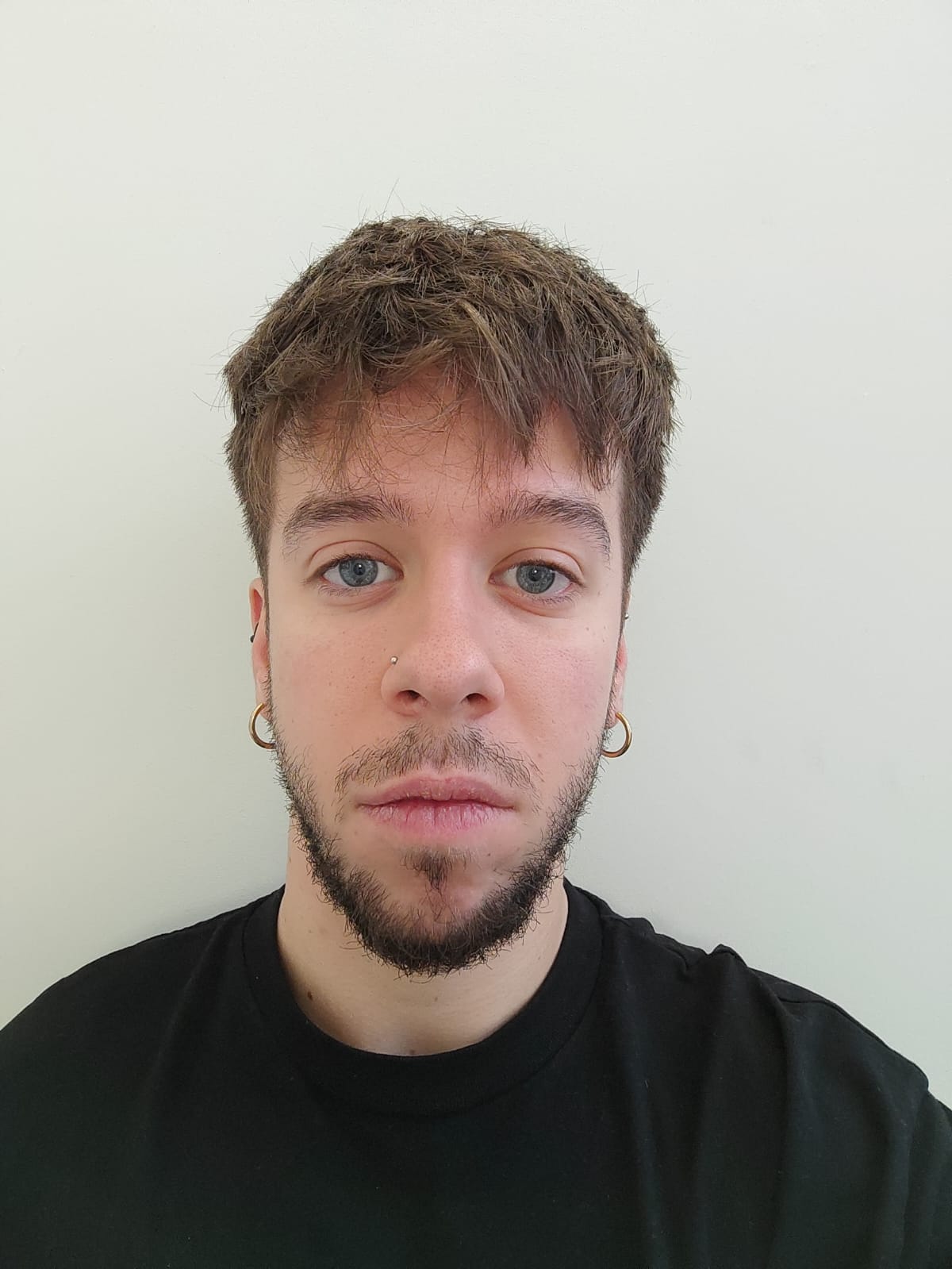}}]{Daniel DeAlcala}, a research professional, obtained his B.Sc. in Telecommunications Engineering from Universidad Autonoma de Madrid (UAM) in 2020, graduating with outstanding academic achievements. Subsequently, he pursued a Master's degree in Deep Learning, which he successfully completed in 2021. In 2022, Daniel embarked on his Ph.D. journey by joining the esteemed Biometrics and Data Pattern Analytics Laboratory (BiDA Lab) at UAM. His research primarily centers around Fair and Transparent AI and innovative architectural developments. Daniel has presented his work at prestigious conferences, including CVPR, and continues to make significant contributions to the field of AI.\end{IEEEbiography}

\begin{IEEEbiography}[{\includegraphics[width=1in,height=1.25in,clip,keepaspectratio]{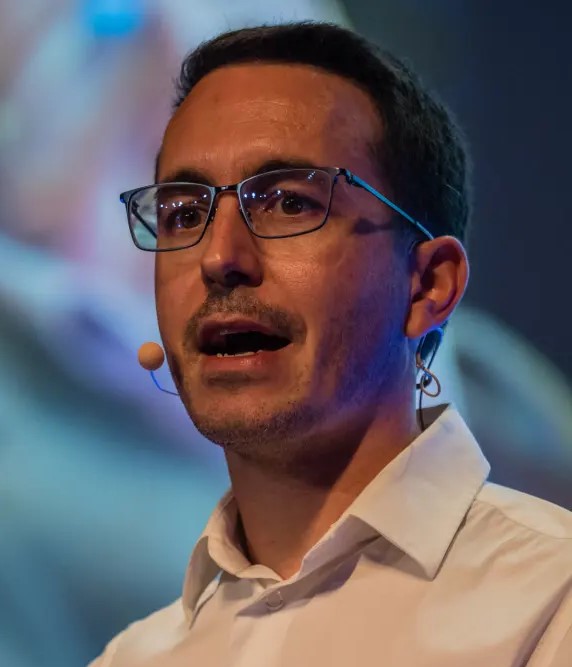}}]{Aythami Morales}
received his M.Sc. degree in Electrical Engineering in 2006 from Universidad de Las Palmas de Gran Canaria. He received his Ph.D. degree in Artificial Intelligence from La Universidad de Las Palmas de Gran Canaria in 2011. He performs his research works in the BiDA Lab – Biometric and Data Pattern Analytics Laboratory at Universidad Autónoma de Madrid, where he is currently an Associate Professor (CAM Lecturer Excellence Program). He is a member of the ELLIS Society (European Laboratory for Learning and Intelligent Systems). He has performed research stays at the Biometric Research Laboratory at Michigan State University, the Biometric Research Center at Hong Kong Polytechnic University, the Biometric System Laboratory at the University of Bologna, and Schepens Eye Research Institute (Harvard Medical School). He is the author of more than 100 scientific articles published in international journals and conferences and 2 patents.\end{IEEEbiography}

\begin{IEEEbiography}[{\includegraphics[width=1in,height=1.25in,clip,keepaspectratio]{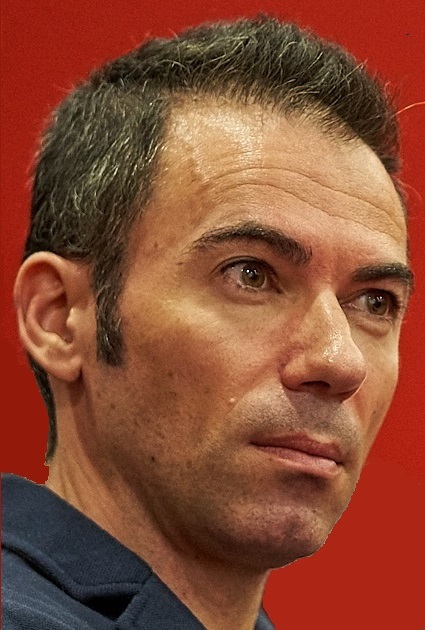}}]{Julian Fierrez}
received the M.Sc. and the Ph.D. degrees in telecommunications engineering from Universidad Politecnica de Madrid, Spain, in 2001 and 2006, respectively. Since 2002 he has been affiliated as a PhD candidate with the Universidad Politecnica de Madrid, and since 2004 at Universidad Autonoma de Madrid, where he is currently a Full Professor since 2022. From 2007 to 2009 he was a visiting researcher at Michigan State University in the USA under a Marie Curie fellowship. Since 2016 he has been Associate Editor for Elsevier's Information Fusion and IEEE Trans. on Information Forensics and Security, and since 2018 also for IEEE Trans. on Image Processing. He has been the General Chair of the IAPR Iberoamerican Congress on Pattern Recognition (CIARP 2018) and the Iberian Conference on Pattern Recognition and Image Analysis (IbPRIA 2019). He is also the recipient of a number of world-class research distinctions, including: the EBF European Biometric Industry Award 2006, EURASIP Best PhD Award 2012, Medal in the Young Researcher Awards 2015 by the Spanish Royal Academy of Engineering, and the Miguel Catalan Award. He is an ELLIS Fellow since 2024. \end{IEEEbiography}

\begin{IEEEbiography}[{\includegraphics[width=1in,height=1.25in,clip,keepaspectratio]{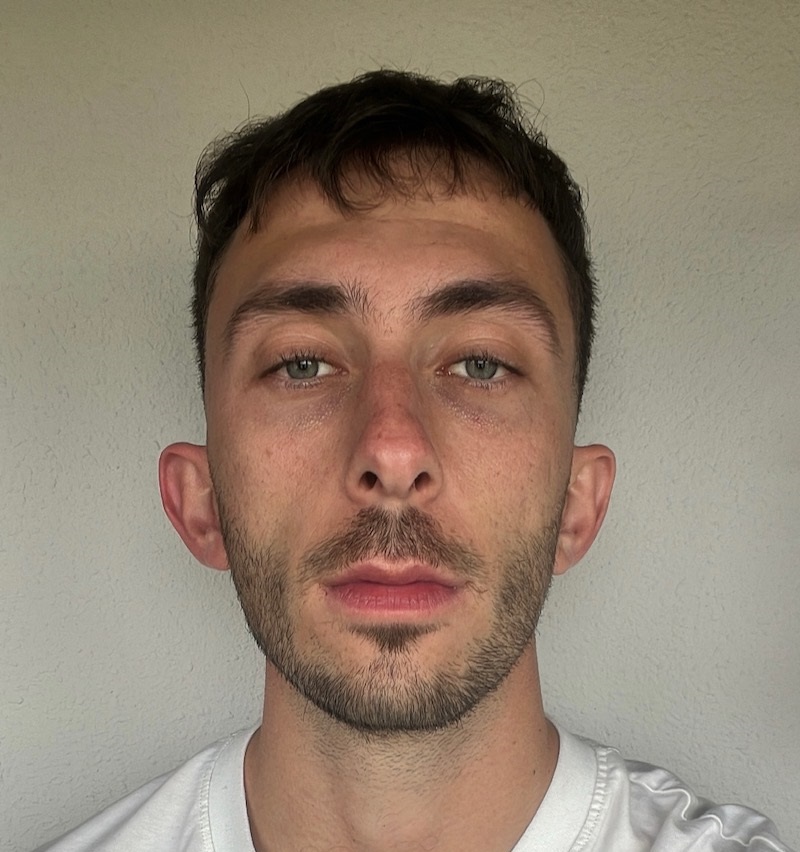}}]{Gonzalo Mancera} earned his B.Sc. in Telecommunications Engineering from Universidad Autónoma de Madrid (UAM) in 2021. Following this, he pursued a Master's degree in Deep Learning fir Audio and Video, which he completed in 2022. In 2023, Gonzalo commenced his Ph.D at the renowned Biometrics and Data Pattern Analytics Laboratory (BiDA Lab) at UAM.
His research initially focused on Fair and Transparent AI and pioneering architectural advancements, and he has since transitioned to working in Natural Language Processing (NLP).\end{IEEEbiography}

\begin{IEEEbiography}[{\includegraphics[width=1in,height=1.25in,clip,keepaspectratio]{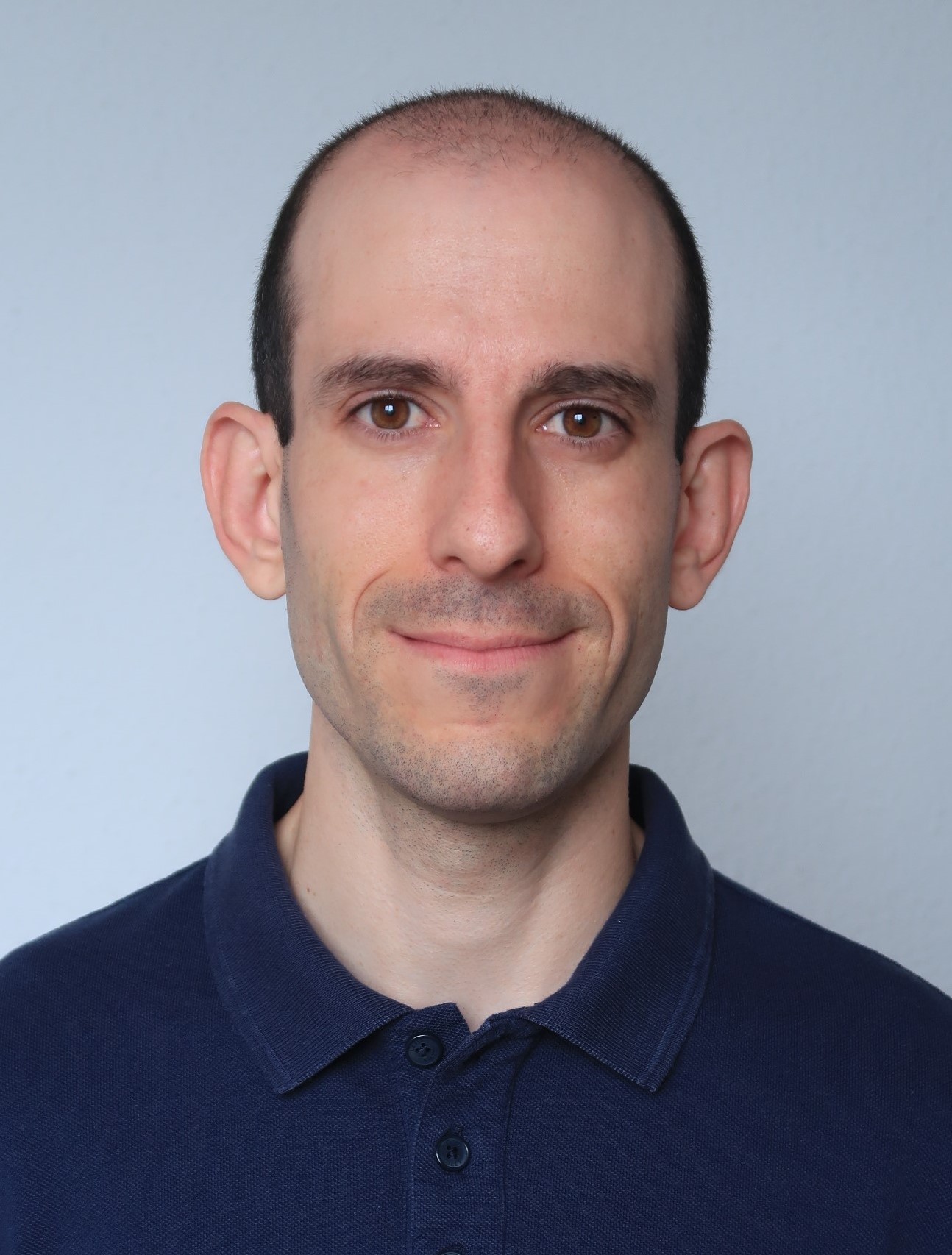}}]{Ruben Tolosana} received the M.Sc. degree in Telecommunication Engineering, and the Ph.D. degree in Computer and Telecommunication Engineering, from Universidad Autonoma de Madrid, in 2014 and 2019, respectively. In 2014, he joined the Biometrics and Data Pattern Analytics - BiDA Lab at Universidad Autonoma de Madrid, where he is currently an Associate Professor. He is a member of the ELLIS Society and ELLIS Unit Madrid, Technical Area Committee of EURASIP, and Editorial Board of the IEEE Biometrics Council Newsletter. His research interests are mainly focused on signal and image processing, pattern recognition, and machine learning, particularly in the areas of DeepFakes, Human-Computer Interaction, Biometrics, and Health. He is the author of over 100 scientific articles published in international journals and conferences in AI, selected as one of the most influential researchers in the world according to “Ranking of the World Scientist: World Top 2\%” in 2023 and 2024, carried out by Stanford University and Elsevier. Dr. Tolosana has also received several awards such as the “European Biometrics Industry Award (2018)” from the European Association for Biometrics (EAB), the “Best Ph.D. Thesis Award in 2019-2022” from the Spanish Association for Pattern Recognition and Image Analysis (AERFAI), and the “Juan Lopez de Peñalver Award (2024)” from the Spanish Royal Academy of Engineering for his research contributions and technology transfer.\end{IEEEbiography}

\begin{IEEEbiography}[{\includegraphics[width=1in,height=1.25in,clip,keepaspectratio]{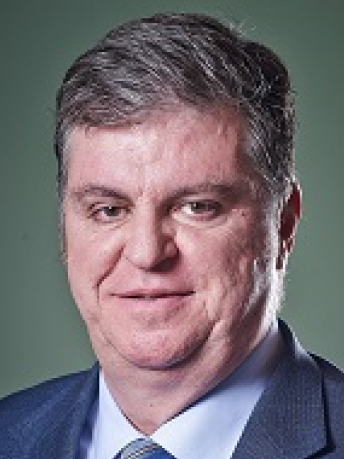}}]{Javier Ortega-Garcia}
received the M.Sc. degree in electrical engineering and the Ph.D. degree (cum laude) in electrical engineering from Universidad Politecnica de Madrid, Spain, in 1989 and 1996, respectively. He is currently a Full Professor at the Signal Processing Chair at Universidad Autonoma de Madrid - Spain, where he holds courses on biometric recognition and digital signal processing. He is a founder and Director of the BiDA-Lab, Biometrics and Data Pattern Analytics Group. He has authored over 300 international contributions, including book chapters, refereed journals, and conference papers. His research interests are focused on biometric pattern recognition (on-line signature verification, speaker recognition, human-device interaction) for security, e-health, and user profiling applications. He chaired Odyssey-04, The Speaker Recognition Workshop, ICB-2013, the 6th IAPR International Conference on Biometrics, and ICCST2017, the 51st IEEE International Carnahan Conference on Security Technology. \end{IEEEbiography}

\appendix
\newpage
\section{Supplementary Material for: Is my Data in your AI Model? Membership Inference Test with Application to Face Images}
\label{sec:AppA}

In this supplementary material, we provide additional results showcasing the application of MINT technology to several image classification models. Specifically, we apply MINT to a range of models, from MobileNet to ViT, using datasets such as CIFAR-100, TinyImageNet, Caltech-256, and Coco.

In the full paper, we introduced the concepts of our MINT proposal and applied them to FR models, where the application of this technology is particularly impactful and engaging for users due to the sensitivity of biometric data, specifically facial images. However, as explained in the paper, the concepts behind our MINT approach are applicable to any type of model or task. This is because it is based on detecting patterns in a model's activations using the so-called MINT Model. In this appendix, we extend this technology to general Image Classification models.

The Image Classification domain provides a more manageable environment where numerous experiments can be conducted. Moreover, training competitive models in this domain is significantly simpler compared to Face Recognition. Results in Tab. \ref{tab:cifarmint}. The experimental protocol follows the same structure as presented in the full paper, with additional details outlined below:

\begin{itemize}
    \item For the $M$ models, we will use the following architectures: MobileNet \cite{howard2017mobilenets}, ResNet50 \cite{he2016deep}, ResNet100 \cite{he2016deep}, DenseNet121 \cite{huang2017densely}, DenseNet201 \cite{huang2017densely}, Xception \cite{chollet2017xception}, and ViT \cite{alexey2020image}. 
    \item The datasets include CIFAR-100 \cite{krizhevsky2009learning}, Tiny-Imagenet \cite{chrabaszcz2017downsampled}, Coco \cite{lin2014microsoft}, and Caltech256 \cite{griffin2007caltech}. The presented models are trained on CIFAR-100 for the Image Classification task, with external data $\mathcal{E}$ sourced from Tiny-Imagenet, Coco, and Caltech256. Tiny-Imagenet and Coco are used as $\mathcal{E}$ for training the MINT Model $T$, while Caltech256 serves as evaluation data.
    \item The selected models and datasets represent some of the most commonly used architectures and benchmarks in the state of the art. 
    \item To train the MINT Model $T$, we use $40,000$ samples from CIFAR-100 utilized to train the Audited Model $M$, representing the training data $\mathcal{D}$, alongside $40,000$ external samples $\mathcal{E}$ from external training datasets (Tiny-Imagenet and Coco). The additional $10,000$ samples from CIFAR-100 are used to evaluate these MINT Models, combined with $10,000$ samples from the external evaluation dataset, in this case, Caltech256.
    \item For the MINT Model $T$, we will use the configuration that achieved the best results during the experiments presented in the full paper. Specifically, we employ a CNN architecture for the MINT Model, using as AAD a convolutional output close to the input of model $M$.
\end{itemize}


\begin{table}[t]
\begin{center}
\caption{Classification accuracy for the CNN MINT Model (Conv Layer \#1 as Auditable Data) trained with $80$K for different state-of-the-art architectures.}
\label{tab:cifarmint}
\resizebox{\columnwidth}{!}{%
\begin{tabular}{@{}ccccccc@{}}
\hline
  MobileNet & ResNet50 & ResNet100 & DenseNet121 & DenseNet201 & Xception & ViT \\ \hline
     0.98     &   0.94   &      0.94     &     0.99        &    0.99         &    0.99      &   0.94    \\ \hline
     0.98     &   0.94   &      0.94     &     0.99        &    0.99         &    0.99      &   0.94    \\ \hline
\end{tabular}}%
\end{center}
\end{table}

\vspace{-10 mm}
\begin{table}[H]
\begin{center}
\caption{Classification accuracy for the CNN MINT Model (Conv Layer \#1 as Auditable Data) trained with $80$K samples for a ResNet50, varying the resolution of the input data.}
\label{tab:cifarmintresolution}
\vspace{5 mm}
\begin{tabular}{@{}lccccccc@{}}
\hline
Resolution & ResNet50 \\ \hline
$16\times16$   &   0.56    \\ \hline
$32\times32$   &   0.64    \\ \hline
$64\times64$   &   0.94    \\ \hline
$128\times128$ &   0.95    \\ \hline
\end{tabular}%
\end{center}
\end{table}

When we train the models ourselves, it is also possible to analyze additional factors. In this case, we will examine the effect of input image resolution on the results. As mentioned in \cite{shafran2021membership}, lower input image resolution negatively impacts the performance of MIAs. Here, we aim to validate this behavior in MINT. Table \ref{tab:cifarmintresolution} presents MINT performance for a ResNet50 model, varying the resolution of CIFAR-100 images prior to training the Audited Model $M$.

\end{document}